\documentclass[12pt]{iopart}
\usepackage{graphicx}
\usepackage{color}
\usepackage{subfigure}

\usepackage{soul}

\begin{document}
\title
{AQPDBJUT Dataset: Picture-Based PM Monitoring in the Campus of BJUT}
\author{Yonghui Zhang$^{1-4}$, Ke Gu$^{1-4}$}
\address{
$^1$Engineering Research Center of Intelligent Perception and Autonomous Control, \tiny\textcolor{white}{A}\small Ministry of Education\\
$^2$Beijing Key Laboratory of Computational Intelligence and Intelligent System\\
$^3$Beijing Artificial Intelligence Institute\\
$^4$Faculty of Information Technology, Beijing University of Technology, China}
\ead{guke.doctor@gmail.com}
\vspace{10pt}
\begin{indented}
\item[]November 2019
\end{indented}

\begin{abstract}
Ensuring the students in good physical levels is imperative for their future health. In recent years, the continually growing concentration of Particulate Matter (PM) has done increasingly serious harm to student health. Hence, it is highly required to prevent and control PM concentrations in the campus. As the source of PM prevention and control, developing a good model for PM monitoring is extremely urgent and has posed a big challenge.
It has been found in prior works that photo-based methods are available for PM monitoring. To verify the effectiveness of existing PM monitoring methods in the campus, we establish a new dataset which includes 1,500 photos collected in the Beijing University of Technology. Experiments show that stated-of-the-art methods are far from ideal for PM monitoring in the campus.
\end{abstract}

\section{Introduction}
Recent years have witnessed the extreme growth of Particulate Matter (PM), leading to an increasing amount of atmospheric environment pollution \cite{01}. PM has become one of the most important factors which affect people's health. It is worth noting that high-concentration PM does potential and permanent harm to student health \cite{02}-\cite{03}. In \cite{02}, Feizabad \emph{et al.} found that the concentration of PM shows a positive association with vitamin D deficiency and a negative association with bone turnover, which indicates that the bones of students who live in a high-concentration PM area for a long time grow much more slowly than their peers. In \cite{03}, Gauderman \emph{et al.} showed that the high-concentration PM is associated with the impairment of lung function between the ages of 10 and 18. As seen, it is urgent to control PM concentration through the real-time PM monitoring data, towards ensuring student health. Relevant researches have received wide concerns from the public during the past few years \cite{3c}-\cite{3g}.

\begin{figure}[!t]
\vspace{0.2cm}
\centering
\subfigure[]{\label{fig1:subfig:a}\includegraphics[width=5.5cm]{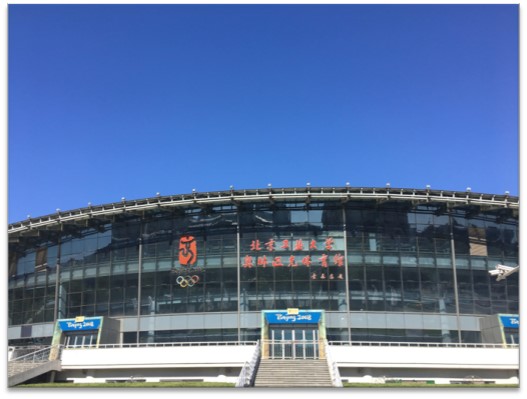}}
\subfigure[]{\label{fig2:subfig:b}\includegraphics[width=5.5cm]{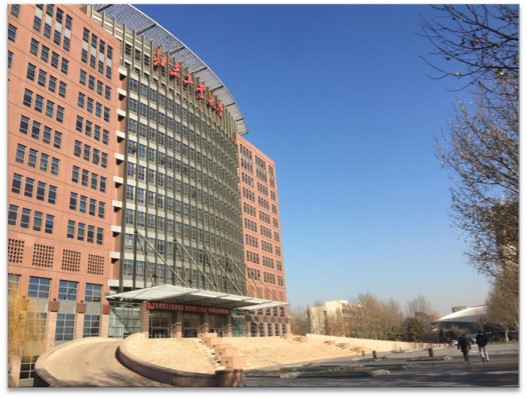}}\\
\subfigure[]{\label{fig1:subfig:a}\includegraphics[width=5.5cm]{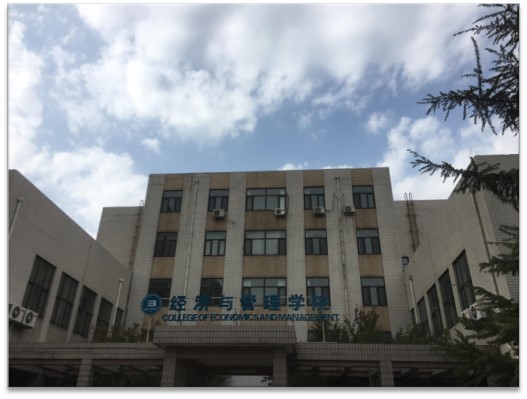}}
\subfigure[]{\label{fig2:subfig:b}\includegraphics[width=5.5cm]{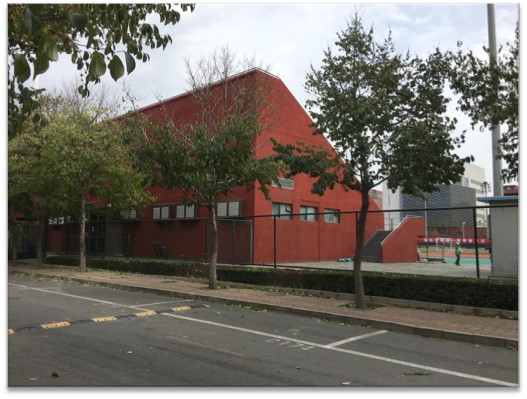}}\\
\flushleft{\small \textbf{Figure 1.} Typical photos in the AQPDBJUT dataset: (a) Science building; (b) Olympic stadium-badminton hall; (c) College of economic and management; (d) Back of playground.}
\label{fig:1}
\vspace{0.4cm}
\end{figure}

To create a good living environment for students, we studied the characteristics of PM monitoring in the campus. Combined with prior studies, it was found that there is a correlation between PM concentration and photos \cite{9a}-\cite{11}. For a future investigation, we establish a new dataset that consists of 1,500 photos taken in the Beijing University of Technology. We called it the AQPDBJUT dataset. The performance of nine state-of-the-art \cite{10}-\cite{05} are examined. Experiments show that their performances are not well.

\section{Dataset}
\begin{figure}
\vspace{0.2cm}
\includegraphics[width=14cm]{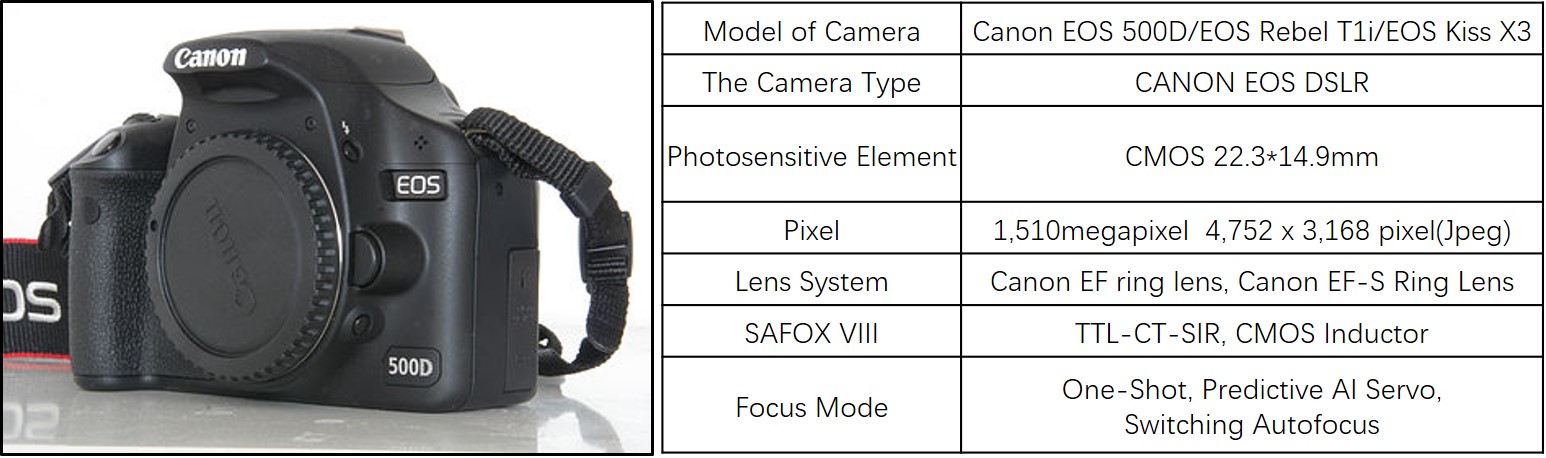}\\
\centering{\small \textbf{Figure 2. }\small The configuration of Canon EOS 500D single-lens reflex camera.}
\label{fig:2}
\end{figure}

The AQPDBJUT dataset is composed of a total number of 1,500 photos of resolution 4,032$\times$3,024. Different from the existing datasets, the photos in the AQPDBJUT dataset were just taken in the Beijing University of Technology (BJUT). The equipment used is Canon EOS 500D, a single-lens reflex camera as shown in Fig. \ref{fig:2}. In this dataset, the photos were captured in different seasons and times over the past three years. It has the characteristics of strong coverage, high definition, etc. Specifically, these photos contain relatively limited scenes, mainly including teaching buildings, playgrounds, trails, and so forth, around student life trajectory. We appropriately increased the number of photos in the locations which are the high-frequency sites for student's outdoor life. This makes the AQPDBJUT dataset more suitable for PM monitoring in the campus.

\begin{figure}
\vspace{0.6cm}
\includegraphics[width=14cm]{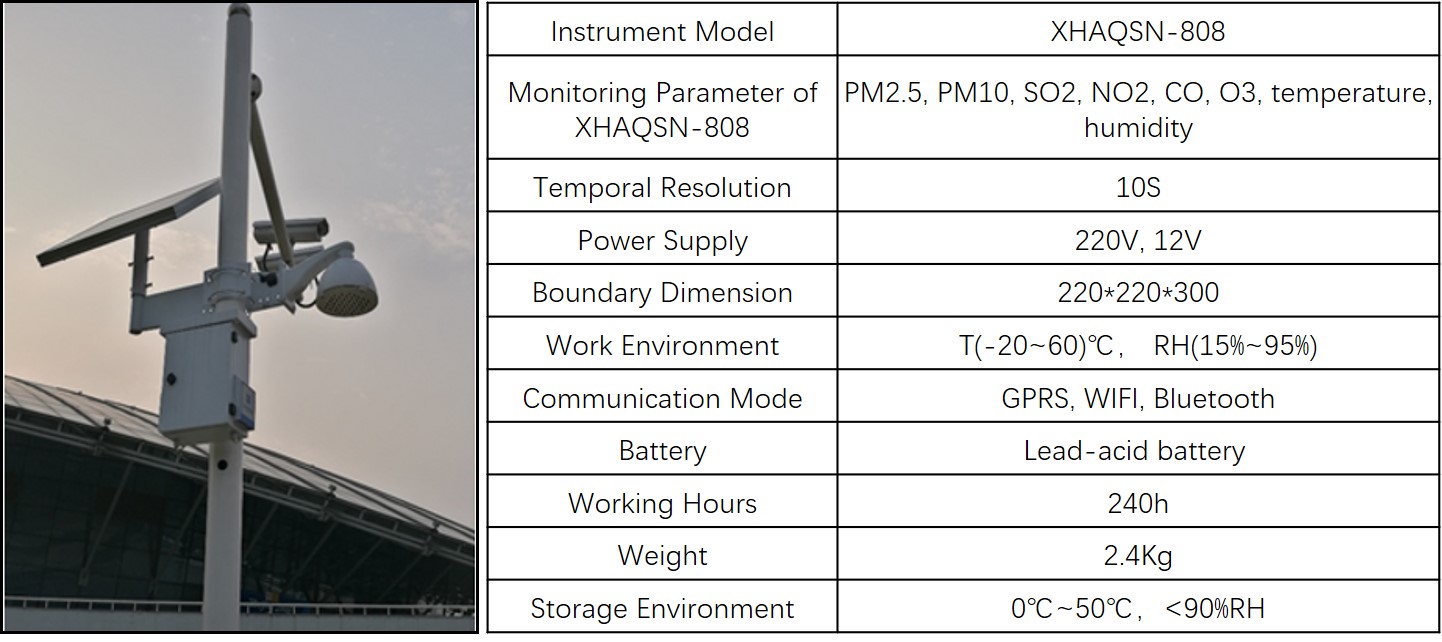}\\
\centering{\small \textbf{Figure 3. }\small Sensor-based real-time PM monitoring equipment `XHAQSN-808'.}
\label{fig:3}
\vspace{0.2cm}
\end{figure}

A professional PM monitoring device called `XHAQSN-808' has been equipped in the campus of Beijing University of Technology. Its detailed parameters are illustrated in Fig. \ref{fig:3}. Base on that device, the more accurate and real-time monitoring data can be obtained to assign the photos. So, the photos in the AQPDBJUT dataset can better reflect the situation of students exposed to high-concentrations PM. According to the statistics of our monitoring device, the real-time monitoring of PM concentration in the AQPDBJUT dataset spans up to 0-350 $\mu g/m^3$.

\section{Experiment}
We calculated the performance nine state-of-the-art models on the AQPDBJUT dataset. We choose three typical criteria to evaluate the model's monitoring performance, including the Root Mean Square Error (RMSE), the Normalized Mean Gross Error (NMGE), and the error-sensitive Peak Signal to Noise Ratio (PSNR). The photos quality methods include NIQMC \cite{04}, BIQME \cite{05}, FISH \cite{06}, FISHBB \cite{06}, ARISM \cite{07}, NIQE \cite{08}, ASIQE \cite{09}, PPPC \cite{10}, and GSWD \cite{11}. A good model is expected to obtain low values in RMSE and NMGE, but high value in PSNR. The result of the above ten models are shown in Table 1. It is not difficult to find from the table that the model FISHBB has obtained the best performance in the PM$_{2.5}$ concentration.

\begin{figure}
\vspace{0.15cm}
\centering{\small \textbf{Table 1.}Comparison of PM$_{2.5}$ concentrations between nine state-of-the-art\\ methods on the AQPDBJUT dataset.}\\
\vspace{0.3cm}
\centering
\label{fig2:subfig:b}\includegraphics[width=11cm]{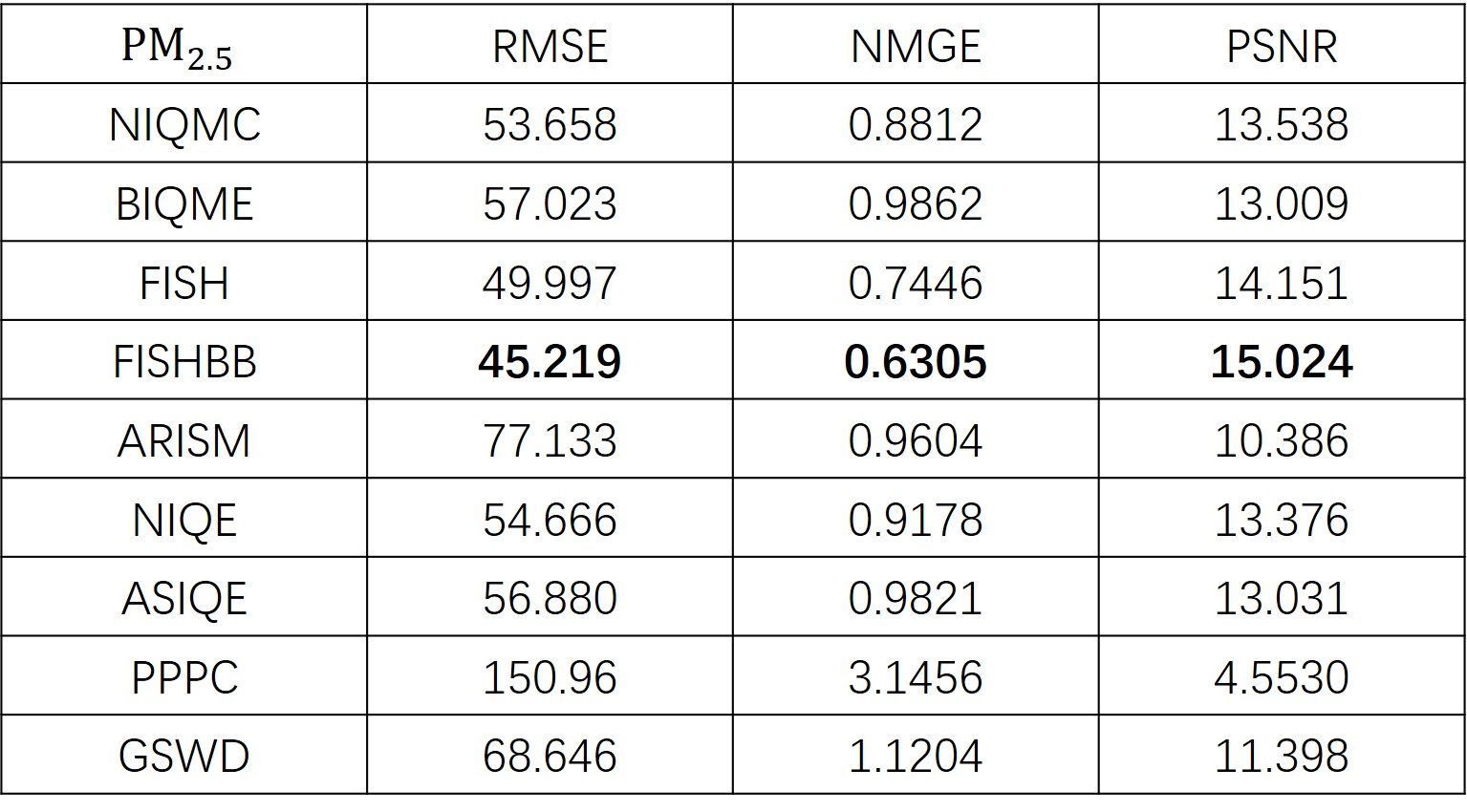}
\label{fig:4}
\end{figure}

\section{Conclusion}
With the rapid development of economic, more and more attention has been concentrated on students health. However, at present, the monitoring system of PM concentration in the campus is still lower, which seriously affects the following governance and prevention. In order to facilitate PM monitoring, we first established a new dataset called AQPDBJUT, in which all the photos were captured in the Beijing University of Technology. Then, we selected nine state-of-the-art models to make an experiment. Experiment shows that nine state-of-the-art methods are not perfect in the AQPDBJUT dataset and cannot accurately monitor the PM concentration in the campus.

\end{document}